\documentclass[11pt]{article}
\usepackage[a4paper,margin=25mm]{geometry}
\usepackage[T1]{fontenc}
\usepackage[utf8]{inputenc}
\usepackage{lmodern,microtype}
\usepackage{amsmath,amssymb,graphicx,booktabs,array,placeins,float,caption,hyperref}
\hypersetup{colorlinks=true,linkcolor=blue,citecolor=blue,urlcolor=blue,
pdftitle={The Conflict Between Logic and Memory: Training Conditions for Optimizer-Dependent Rule Acquisition},pdfauthor={Gongyue Zhang and Honghai Liu}}
\title{\textbf{The Conflict Between Logic and Memory:}\\Training Conditions for Optimizer-Dependent Rule Acquisition}
\author{Gongyue Zhang \and Honghai Liu}
\date{October 2, 2026}
\begin{document}
\maketitle
\begin{abstract}
Optimizers can fit the same task while acquiring different generalizing relations. We study the training conditions governing these differences in single-hidden-layer ReLU networks, combining composite evidence tasks, parameter-level interventions, and a three-seed strict-parity scan. Our central finding is that nuisance-connected trainability reshapes both shared failures and relative optimizer advantages. In a nuisance-heavy task, all twenty tested optimizer configurations remain near chance on the hardest stage. Retaining every input but fixing nuisance-connected first-layer weights at initialization raises that stage's accuracy from approximately 50\% to 70.56\%, 68.47\%, and 69.00\% for momentum SGD, Adam, and Muon. Masking the same inputs only after full training does not recover this performance. On a separate pairwise-mode task, background freezing reduces Muon's rare-mode advantage over momentum SGD by 12.48 percentage points, while the target and mode frequencies remain fixed. Each intervention is evaluated under a common validation-selection protocol with condition-specific learning rates and checkpoints. A strict-parity sweep over orders 1--20 provides a complementary reference without spurious cues or extra nuisance coordinates: the optimizers separate at orders 9--11, then approach chance despite substantial remaining Bayes predictability. A mixed task establishes a recovery boundary, and CIFAR-10 supplies an external architecture comparison. Together, these findings connect optimizer comparison to the acquisition and use of specified relations, identifying permitted adaptation as a concrete training variable that changes what a fixed architecture learns.
\end{abstract}

\section{Introduction}
\label{sec:introduction}

Success on familiar examples does not always transfer to new examples governed by the same relation. Everyday comparisons between humanities and science learning, including those common in Chinese education, make this distinction intuitive: reproducing material and using a rule in a new situation place different demands on a learner. The scientific question concerns the learning process rather than a division of people into cognitive types: what determines whether fitting experience yields reusable knowledge?

Evidence from cognition and neuroscience motivates attention to how learning is regulated. In a statistical-learning task, Sherman and Turk-Browne showed that predicting upcoming information can interfere with episodic encoding of the present~\cite{sherman2020}. Causal manipulations of astrocyte activity can also alter neuronal representations supporting memory and associated behavior~\cite{bukalo2026}. These results concern different biological processes, but both motivate studying the conditions under which a system acquires and expresses information. In artificial networks, optimizers provide a controllable entry point: a fixed architecture can develop different predictive functions under different parameter-update rules.

Research on implicit bias has made this dependence concrete. Adaptive and nonadaptive methods can generalize differently~\cite{wilson2017}; recent work connects Adam to richer feature learning~\cite{vasudeva2025} and Muon's spectral updates to more balanced learning under data imbalance~\cite{vasudeva2026}. These explanations raise a further question: how do the resulting functions use the particular relations that make a prediction correct? Feature diversity, balanced component learning, and acquisition of a specified target relation measure different aspects of learning. We study the following question:
\begin{quote}
\emph{Which reusable target relations do different optimizers learn to use, and which training conditions determine whether their observed advantages persist?}
\end{quote}

We investigate this question in one-hidden-layer ReLU networks with known data-generating relations. Throughout, \emph{memory} denotes sample-specific fitting, and \emph{logic} denotes generalization of reusable target relations; these are operational terms, not claims about separate neuronal systems. We distinguish true predictive inputs, spurious cues whose training correlation does not persist at test time, and nuisance inputs that are irrelevant to the target by construction. The experiments combine composite tasks with dispersed evidence, competing cues, rare relational modes, and irrelevant inputs, together with a strict parity family. We compare momentum SGD (SGDM), Adam, and Muon, and extend the composite comparison to 20 optimizer configurations. Within each task, paired data, initializations, and training schedules support comparisons at the level of stages, modes, and cue dependence.

The resulting differences are not captured by an overall ranking. In the rare-mode task, Muon's advantage over SGDM is about 1 percentage point overall but 12 points on the rare mode. Conversely, expanding the optimizer set changes overall performance without resolving the hardest stage of the nuisance-heavy composite task: all 20 selected configurations achieve only 49.86\%--50.53\% fourth-stage accuracy. These observations motivate a task-internal intervention: determine whether an apparent limitation or advantage survives a change in which input connections are allowed to adapt.

Our central finding is that nuisance-connected trainability reshapes both shared failures and relative optimizer advantages. We retain all inputs and fix selected first-layer weights at their initialization, so nuisance inputs continue to enter the network through their initial random projections. In the nuisance-heavy task, fourth-stage accuracy rises from approximately 50\% to 70.6\%, 68.5\%, and 69.0\% for SGDM, Adam, and Muon, respectively. Masking the same nuisance inputs only after full training leaves accuracy near chance. In the rare-mode task, freezing background connections changes SGDM/Adam/Muon rare-mode accuracy from 55.6/56.0/67.6\% to 84.9/88.6/84.4\%. The target rule and mode frequencies remain unchanged, yet Muon's approximately 12-point advantage over SGDM disappears. Its reduction is 12.2--13.0 points in each of the three reporting seeds. Thus the intervention identifies both a shared obstacle and a condition governing an optimizer comparison. These are effects under a common validation-selection protocol, with learning rates and checkpoints selected separately for each condition; an earlier fixed-learning-rate, fixed-epoch experiment provides a complementary control.

Strict parity addresses a complementary question: how far does training progress when the target offers no lower-order predictive entry point? Across orders 1--20, three seeds, and a fixed $32\!\to\!64\!\to\!2$ architecture, SGDM, Adam, and Muon all attain high accuracy through order eight, then separate at orders nine to eleven. At order ten their accuracies are 84.78\%, 78.14\%, and 74.96\%; at order eleven they are 63.44\%, 55.84\%, and 52.25\%. At orders thirteen to twenty all approach chance, while the Bayes reference remains approximately 99.1\%--90.5\%. The scan characterizes optimizer-dependent learning under a finite budget while distinguishing available statistical information from achieved prediction. The composite interventions also expose a persistent failure: removing all designated non-target inputs in the mixed task leaves fourth-stage performance near chance. Recovery therefore has a measurable boundary within the same experimental framework.

Our contributions are threefold:
\begin{enumerate}
\item \textbf{Nuisance-connected trainability reshapes rule generalization and optimizer advantages.} Keeping nuisance exposure intact while restricting adaptation recovers a shared failure and removes a rare-mode advantage. Test-time masking does not reproduce the former recovery, locating the effective intervention in training.
\item \textbf{Overall optimizer rankings and relation- or mode-level performance can diverge.} Across structured tasks and an expanded optimizer comparison, we show where aggregate gains coexist with unresolved stage-level failure, and where aggregate scores understate a substantial rare-mode improvement.
\item \textbf{Finite-budget higher-order learning profiles across update rules.} A three-seed strict-parity scan identifies a region of optimizer separation and a subsequent shared loss of generalization despite substantial remaining predictive information.
\end{enumerate}

Together, these results connect optimizer comparison to the acquisition and use of specified relations. They provide experimentally testable conditions for explaining why learning succeeds, fails, or changes its relative advantage within the same shallow architecture. Section~2 situates these questions in prior work; Section~3 defines the tasks, update constraints, and functional measurements; Section~4 reports the evidence; Section~5 concludes with their implications for optimization.

\section{Related Work}
\label{sec:related}

\subsection{Optimizer bias and the acquisition of specified relations}

Implicit-bias research establishes that optimization can select different predictors even when training performance is comparable. Wilson et al.~\cite{wilson2017} demonstrated consequential generalization differences between adaptive methods and SGD. Simplicity-bias studies show how an easily learned predictor can dominate more reliable features~\cite{shah2020}. Beneventano et al.~\cite{beneventano2024} connect SGD's stochasticity to learning relevant input support and suppressing dependence on irrelevant coordinates. These results make the learned function, including which inputs it uses, central to optimizer comparison.

\emph{The Rich and the Simple}~\cite{vasudeva2025} is a direct neighbor. It links Adam to richer features and decision boundaries, and evaluates both robustness and decoded core-function information. Its Boolean experiment places a first-order spurious cue against a thresholded staircase core containing $x_1,x_1x_2,\ldots$. This structure has a low-order entry point; it differs from strict parity, where every proper subset of the independent latent factors is uninformative about the label. Our comparison adds this distinction and tests whether optimizer advantages persist when nuisance-connected trainability changes.

Vasudeva et al.~\cite{vasudeva2026} analyze how Muon's spectral design balances component learning in linear and deep-linear settings, with experiments on imbalanced classification. Our rare-mode experiment asks a task-internal question: does an observed advantage persist when the mode and its frequency remain fixed but background connections cannot adapt? Its disappearance identifies an additional condition that an explanation of the measured advantage must accommodate. Background-adaptation effects can coexist with spectral balancing. These comparisons extend the study of feature richness and component balance to the conditions under which specified target relations guide prediction.

\subsection{Feature competition, spurious cues, and nuisance fitting}

Feature competition supplies a basis for understanding why fitting and relation generalization can separate. Gradient starvation describes how learning one predictive feature can inhibit learning another~\cite{pezeshki2021}. Qiu et al.~\cite{qiu2024} study how core and spurious function complexity affects feature learning, including parity and staircase constructions. LaBonte and Muthukumar~\cite{labonte2026} analyze SGD's prioritization of a shortcut in an XOR model. These studies already establish competition between strict higher-order targets and first-order shortcuts.

Nuisance fitting creates a related challenge even without a predictive shortcut. Analyses of benign overfitting in two-layer convolutional networks track the development of signal and noise components~\cite{cao2022}, including XOR data~\cite{meng2024}. Huang et al.~\cite{huang2025} study how label-noise gradient descent can improve generalization in a low-signal-to-noise regime. These works demonstrate that training success, signal learning, and noise fitting need not advance together; they also show why the form and location of noise matter. Label corruption, noisy observations of a true factor, and additional target-irrelevant coordinates are distinct interventions.

Our experiments connect these competition questions to comparisons among update rules. Cue reversal tests reliance on a training correlation; stage- and mode-specific evaluation locates the remaining failures; training interventions test how those outcomes change. We find both a nuisance-heavy task in which several optimizers share a recoverable failure and a rare-mode task in which their responses differ enough to remove the original gap. A mixed task supplies the complementary case in which designated nuisance removal is insufficient. The resulting object of study is the pattern of shared and optimizer-dependent responses, not merely the existence of competition or noise memorization.

\subsection{Training interventions, representations, and readout}

Prior interventions distinguish acquiring a representation from using it in the final prediction. Deep Feature Reweighting shows that retraining the last layer can recover subgroup robustness from representations learned with spurious correlations~\cite{kirichenko2023}. Freeze then Train preserves representations learned in an unsupervised stage before subsequent supervised training~\cite{ye2023}. Feature Sieve uses an intervention to counter simplicity bias and encourage learning beyond easy predictive features~\cite{tiwari2023}. These approaches establish that training constraints and subsequent readout can change which information determines a model's behavior.

Our intervention isolates three experimentally distinct operations: exposing the network to nuisance inputs, allowing their first-layer connections to adapt, and retaining their contribution at inference. Training-time zeroing removes the inputs; freezing preserves their initial projections but prevents the selected weights from changing; test-time masking alters an already trained predictor. Comparing these conditions within each optimizer identifies effects of training restrictions that inference-time cleaning does not reproduce. The intervention uses generator-known groups as a diagnostic control, and does not presuppose an algorithm that discovers irrelevant connections from unannotated data.

This design identifies a control variable and its functional consequences, while leaving the internal route open. Freezing selected columns changes their interaction with trainable columns, hidden-unit gates, and the output layer. Failure of post-training masking does not prove that the target information is absent from the learned representation: a different readout could still recover it. Moreover, separately selected learning rates and checkpoints make the main comparison an intervention effect under the stated selection protocol, not an isolated instantaneous effect of one update operator.

\subsection{Higher-order structure, expressivity, and finite-budget learning}

Higher-order learning has a substantial theoretical and empirical foundation. Barak et al.~\cite{barak2022} study hidden progress in learning parity with SGD, while Glasgow~\cite{glasgow2024} analyzes feature discovery and subsequent tuning for XOR. The merged-staircase property~\cite{abbe2022} and leap complexity~\cite{abbe2023} relate learnability to the organization of interactions and the availability of intermediate structure. Cornacchia and Massouli\'e~\cite{cornacchia2026} study shallow networks on cluster-structured features. These works show why maximum polynomial degree alone is an incomplete description of learning difficulty: tasks with similar highest degrees can provide very different routes to predictive information.

We consequently distinguish composite task stages from strict interaction order. Composite targets can expose useful lower-order prefixes, so success on a nominal second or third stage need not require a second- or third-order computation. In the strict family, $y=\prod_{j=1}^{k}z_j$ for independent balanced latent signs; every proper subset of the factors is label-uninformative. Their noisy observations retain the distinction, although increasing $k$ at fixed total signal norm also reduces the signal allocated to each factor. The parity scan therefore measures learning along a precisely specified task family, including its observation noise, rather than a distribution-independent effect of order alone.

Expressivity and successful optimization must also be distinguished~\cite{malach2021}. Bias-free ReLU networks have structural restrictions relevant to parity and symmetry~\cite{zhang2025}; this structural restriction supplies an architectural reference, separate from the trainability interventions. Likewise, a high Bayes reference establishes predictive information in the observation distribution, not that the fixed-width network can attain it. The results apply to the specified implementations, learning-rate searches, and stopping protocol, including the reported search-boundary cases. Within these conditions, the strict-parity scan identifies where update rules separate without a lower-order entry point and where all lose generalization under the tested budget.

\section{Optimizer-Dependent Rule Acquisition}
\label{sec:framework}

The experimental object is the function obtained by training, including which target relations guide its predictions. We hold the forward architecture fixed within a comparison and vary the update rule or the connections permitted to adapt. Task structure specifies the available evidence; functional diagnostics locate successes and failures; training interventions test the conditions on which an optimizer's outcome depends.

\subsection{From parameter updates to acquired behavior}
\label{sec:outcomes}
Let $\mathcal O$ denote an optimizer configuration, $\mathcal C$ a training constraint, $\lambda$ its hyperparameters, and $\omega$ the initialization and minibatch schedule. A training trajectory and its validation-selected predictor are
\begin{equation}
 \{\theta_t\}_{t=0}^{T}=\mathcal A(D_{\rm tr},\mathcal O,\mathcal C,\lambda,\omega),
 \qquad \hat\theta=\mathcal S(D_{\rm val},\{\theta_t\},\Lambda),
 \label{eq:training_map}
\end{equation}
where $\mathcal S$ includes the stated checkpoint and learning-rate selection over $\Lambda$. Thus the reported outcome depends on both a training procedure and its selection protocol. For a binary margin $f_\theta$ and prediction $\hat y_\theta=\operatorname{sign}f_\theta$, define
\begin{equation}
 \widehat F(\theta)=\frac1n\sum_{i=1}^n\mathbf1\{\hat y_\theta(x_i)=y_i\},
 \qquad G_Q(\theta)=\Pr_Q[\hat y_\theta(x)=y].
 \label{eq:fit_generalization}
\end{equation}
Labels are written as $\{-1,+1\}$ in the analysis and encoded as $\{0,1\}$ in the binary implementations. The first quantity measures fitting at a fixed checkpoint; the second measures generalization to a specified evaluation distribution $Q$. Online training accuracy, accumulated while parameters change within an epoch, is a separate recorded statistic.

High fitting, successful prediction from target evidence, and independence from a spurious cue are distinct properties. Strict-parity generalization tests a target with no lower-order predictive subset; subgroup and cue interventions test which evidence controls predictions in composite tasks. We use \emph{rule acquisition} operationally for the resulting generalizing behavior, without assuming that accuracy uniquely identifies an internal representation. The optimization question is which behaviors become attainable in training under each update configuration and constraint.

\subsection{Target evidence, spurious cues, and independent nuisance}
\label{sec:input_roles}
Write $x=(x_T,x_S,x_R,x_B)$, comprising target observations, spurious cues, irrelevant relational inputs, and Gaussian background. The last two form $x_N=(x_R,x_B)$. Nuisance is generated independently of the target factors, labels, and spurious cues. Spurious cues instead have a prescribed label agreement in training, reduced to approximately one half at test time. They are input features, not corrupted target labels. No reported synthetic preset deliberately corrupts the labels.

An observed factor occupies a disjoint coordinate block with a fixed unit direction $u$:
\begin{equation}
 x_{\rm block}=a z u+\epsilon,\qquad z\in\{-1,+1\},\qquad
 \epsilon\sim\mathcal N(0,\sigma^2 I).
 \label{eq:channel}
\end{equation}
Directions are fixed by a code seed and shared across splits. Independent relational inputs encode products of factors unrelated to the target; their declared order describes the encoded product. Gaussian background has no encoded relation. Observation noise $\epsilon$ on a true factor is distinct from both forms of extra nuisance. The generator-known grouping is available to the diagnostic intervention, while ordinary training receives only $(x,y)$.

Population independence does not eliminate sample-specific associations. For $n$ examples, independent background $b_i\sim\mathcal N(0,\sigma_B^2 I_{d_B})$, and signed labels, let $c_B=n^{-1}\sum_i y_i b_i$. Then
\begin{equation}
 \mathbb E[c_B]=0,\qquad \mathbb E\|c_B\|_2^2=\frac{d_B\sigma_B^2}{n}.
 \label{eq:empirical_noise}
\end{equation}
This elementary calculation identifies an opportunity for fitting incidental associations whose total squared size grows with dimension. It does not specify how a particular optimizer uses those associations. That dependence is tested by changing the permitted adaptation to their coordinates.

\subsection{Composite evidence and strict interaction order}
\label{sec:task_families}
\paragraph{Composite tasks.}
Four tasks vary the organization of reliable evidence, cue competition, and nuisance (Table~\ref{tab:task_design}). All use disjoint eight-coordinate views embedded in 3,124 inputs. Their exact amplitudes, noise scales, coordinate counts, and preset identifiers appear in Appendix~\ref{app:reproduction}.

\begin{table}[H]
\centering\small
\caption{Composite task designs. The table specifies the questions, independently of the observed optimizer rankings. All tasks include Gaussian background; its scale varies across presets.}
\label{tab:task_design}
\begin{tabular}{>{\raggedright\arraybackslash}p{0.18\linewidth}>{\raggedright\arraybackslash}p{0.38\linewidth}>{\raggedright\arraybackslash}p{0.35\linewidth}}\toprule
Task & Target and competing inputs & Question\\\midrule
Distributed votes & Majority of 17 independent signs; one concentrated first-order cue & Can distributed valid evidence prevail over a strong shortcut?\\[3pt]
Higher-order nuisance & Staged target with constituent orders 1--4; irrelevant relations and background; no spurious cue & Does adaptation to irrelevant inputs limit difficult-stage learning?\\[3pt]
Rare modes & Three pairwise modes with frequencies 60/30/10; a first-order cue and nuisance & Does a rare-mode advantage survive restricted background adaptation?\\[3pt]
Mixed evidence & Staged target; one spurious relation at each order; relational and background nuisance & When is removing designated interference insufficient?\\\bottomrule
\end{tabular}
\end{table}

The vote label is $y=\operatorname{sign}(\sum_{j=1}^{17}z_j)$ for independent balanced signs. Each evidence source is first order, while the thresholded majority function is not a pure degree-one target. In the mode task, a mode $m$ is drawn with probabilities $(0.6,0.3,0.1)$ and $y=z_{m,1}z_{m,2}$. Only that mode's two factor blocks carry signal; other mode blocks contain observation noise. A single factor is uninformative about $y$, including conditional on its mode. The learner must combine the active pair to use the target evidence.

For the staged targets, three nested chains provide relations $q_{j,c}=\prod_{\ell=1}^{j}z_{\ell,c}$ for $j<K$ and $c=1,2,3$; only chain 1 continues at $K=4$. Define
\begin{align}
 H(z)&=\sum_{j=1}^{K-1}a_j\sum_{c=1}^3q_{j,c}+\tfrac32q_{K,1},
 & y&=\operatorname{sign}H(z),\qquad a_j=2^{K-j-1},\label{eq:staged_target}\\
 J(z)&=\min\{j:|H_{\leq j}(z)|>R_j\},
 & R_j&=\sum_{\ell=j+1}^{K-1}3a_\ell+\tfrac32\quad(j<K),\quad R_K=0.
 \label{eq:stage_definition}
\end{align}
Here $H_{\leq j}$ retains terms through $j$, and $R_j$ bounds the remaining score. The stage $J$ is the first certificate under this prefix procedure. The generator samples specified stage proportions while preserving the original conditional law within each stage. This reweighting can change the joint latent distribution; independence assumptions for strict parity are not imposed on these composite tasks. Stage identifiers are diagnostic metadata and are not model inputs.

Stage is consequently distinct from minimum predictive order or the Fourier degree of the thresholded label. In the four-stage construction, conditional on the diagnostic stage, first- and second-prefix information already predicts second- and third-stage labels, respectively. The first three prefixes leave fourth-stage labels balanced. These properties describe conditional information, not the computation performed by a trained network. Stage-level performance must therefore be interpreted alongside functional interventions; the explicit construction and audit are given in Appendix~\ref{app:stage_audit}.

\paragraph{Strict parity.}
The separate family used in FWSTEP03 takes independent balanced signs $s_1,\ldots,s_k$ and
\begin{equation}
 y=\prod_{j=1}^k s_j,\qquad
 x_j=\alpha_k s_j u_j+\epsilon_j,\qquad
 \alpha_k=\frac4{\sqrt k},\quad\epsilon_j\sim\mathcal N(0,0.35^2 I).
 \label{eq:strict_parity}
\end{equation}
The 32 coordinates are partitioned as evenly as possible into $k$ disjoint factor views. Every clean sample has norm four. No spurious cue, independent relational distractor, or extra background coordinate is included. For every proper subset $S\subsetneq\{1,\ldots,k\}$,
\begin{equation}
 \Pr(y=1\mid x_S)=\tfrac12.
 \label{eq:strict_subset}
\end{equation}
An omitted independent factor keeps parity uniform, even when the retained factors are observed without error. Thus order here specifies a joint information requirement, not a number of layers or a derivative order.

Writing $v_j=u_j^\top x_j$ and $\sigma=0.35$, the known generator gives
\begin{equation}
 p_k(x):=\Pr(y=1\mid x)
 =\frac{1+\prod_{j=1}^{k}\tanh(\alpha_k v_j/\sigma^2)}{2},
 \qquad A_k^\star=\frac{1+(1-2\Phi(-\alpha_k/\sigma))^k}{2}.
 \label{eq:parity_bayes}
\end{equation}
The reported Bayes reference averages $\max\{p_k(x),1-p_k(x)\}$ over test observations. It tracks remaining statistical information as the signal per factor decreases with order; it is distinct from accuracy proved attainable by a fixed-width MLP. Appendix~\ref{app:bayes} derives both expressions.

\subsection{Update rules and exposure--adaptation interventions}
\label{sec:update_interventions}
\paragraph{Implemented update configurations.}
Synthetic SGDM uses momentum $0.9$ without Nesterov acceleration:
\begin{equation}
 v_t=0.9v_{t-1}+g_t,\qquad \theta_{t+1}=\theta_t-\eta v_t,
 \label{eq:sgdm}
\end{equation}
with zero initial velocity. Adam~\cite{kingma2015} uses
\begin{equation}
 m_t=\beta_1m_{t-1}+(1-\beta_1)g_t,\quad
 v_t=\beta_2v_{t-1}+(1-\beta_2)g_t^2,\quad
 \Delta\theta_t=-\eta\frac{\hat m_t}{\sqrt{\hat v_t}+\varepsilon},
 \label{eq:adam}
\end{equation}
where hats denote bias correction, $(\beta_1,\beta_2)=(0.9,0.999)$, and $\varepsilon=10^{-8}$. Products, squares, and division are coordinatewise. Muon applies a momentum-based, approximately orthogonalized matrix update~\cite{jordan2024}: the implementation uses momentum $0.95$, Nesterov enabled, and five Newton--Schulz steps on matrix parameters, including the output matrix. Vector parameters use its AdamW fallback with $(0.9,0.95)$ and $\varepsilon=10^{-10}$; the adjusted-learning-rate option is disabled. All synthetic configurations have zero weight decay. These settings define the compared procedures; the name Muon includes its vector fallback.

The three rules process coordinate magnitudes and matrix directions differently, creating testable differences in adaptation. Adam's squared-gradient accumulator is not a Hessian, and none of these update formulas assigns a fixed logical order to an optimizer. We examine their acquired functions and intervention responses instead.

\paragraph{Training constraints.}
For a selected input block $C$, let $M_C$ zero its first-layer gradient columns and let $\Pi_C$ restore the corresponding weights to their initialization. A frozen trajectory uses full inputs and performs
\begin{equation}
 (\tilde\theta_{t+1},s_{t+1})=\mathcal U_{\mathcal O}(\theta_t,s_t,M_Cg_t;\eta),
 \qquad\theta_{t+1}=\Pi_C(\tilde\theta_{t+1}),
 \label{eq:constrained_update}
\end{equation}
where $s_t$ is optimizer state. Restoration after every step enforces $W_{:,C}(t)=W_{:,C}(0)$, including for matrix-coupled updates. Gradient masking occurs before the optimizer's transformation, so the transformation of unmasked directions may also change. The intervention is this complete constrained training procedure.

\begin{table}[H]
\centering\small
\caption{Training and diagnostic operations for block $C$. ``Full'' means the original inputs, including $C$. Each training condition selects its own learning rate and checkpoint. Diagnostics modify only an already selected model.}
\label{tab:operations}
\begin{tabular}{p{0.22\linewidth}p{0.22\linewidth}p{0.22\linewidth}p{0.22\linewidth}}\toprule
Operation & Training input & Selected columns & Primary test input\\\midrule
Full training & Full & Adaptive & Full\\
Training freeze & Full & Fixed at $W_{:,C}(0)$ & Full\\
Training mask & $x_C=0$ & Optimizer retained & $x_C=0$\\
Test-only mask & Full & Fully trained & $x_C=0$\\
Test-only reset & Full & Reset after training & Full\\\bottomrule
\end{tabular}
\end{table}

Freezing retains the contribution $W_{:,C}(0)x_C$ to hidden preactivations. Other parameters, including the output layer, can still adapt to that contribution. It preserves the forward architecture but reduces the admissible parameter trajectories and the number of trainable weights. Training masking changes exposure and the validation distribution; the same mask-trained checkpoint is also tested with full inputs restored. Test-only masking changes the input to a fully trained function. The recorded reset diagnostic instead restores selected columns after full training. Its selected columns then match those in a frozen model, while the rest of the network has followed a different history. Main conclusions below use the masking and freezing comparisons; reset records remain available for additional diagnosis.

\subsection{Functional measurements and optimizer--constraint comparisons}
\label{sec:functional_measures}
For a generator-defined group $g$, report $A_g=\Pr[\hat y=y\mid g]$, using stages for the staged targets and modes for the pairwise task. Overall accuracy weights these groups by their frequencies and can conceal a shared hard-stage failure or a large rare-mode gap. Cross-entropy accompanies accuracy to distinguish correct decisions from predictive probability quality.

For cue sign $s$, matching and conflicting accuracies condition on $s=y$ and $s\neq y$. The reflection diagnostic changes one observed factor of a cue relation:
\begin{equation}
 \mathcal R_u(x_{\rm block})=x_{\rm block}-2u(u^\top x_{\rm block}),
 \qquad R_{\rm flip}=\Pr[\hat y(x)\neq\hat y(\mathcal R_Sx)].
 \label{eq:cue_reflection}
\end{equation}
Reflection reverses the encoded sign while preserving the isotropic Gaussian channel law. For a multi-factor cue, reflecting one factor reverses its product. The vote and mode diagnostics each concern a single cue. Low flip sensitivity is meaningful together with high target accuracy; a constant predictor would also have zero sensitivity.

To express the optimizer-side question directly, let $A_{g,o}^{C}$ be group accuracy for optimizer $o$ under constraint $C$ and the specified selection protocol. Define
\begin{equation}
 \Delta_{g,o}^{C}=A_{g,o}^{C}-A_{g,o}^{\rm full},\qquad
 \Gamma_{g;o_1,o_2}^{C}
 =(A_{g,o_1}^{C}-A_{g,o_2}^{C})-(A_{g,o_1}^{\rm full}-A_{g,o_2}^{\rm full}).
 \label{eq:optimizer_interaction}
\end{equation}
The first contrast asks whether restricting adaptation improves one optimizer's outcome. The second asks whether that restriction changes the gap between two optimizers. It is a descriptive paired contrast over seeds, not an asserted internal mediation decomposition. This distinction connects the common recovery in Section~\ref{sec:fw_intervention} to the changing rare-mode advantage in Section~\ref{sec:fw_conditional}. Each is an optimization finding: the first identifies a shared dependence on permitted adaptation; the second identifies a condition governing a relative advantage.

\subsection{Architectural capacity and access to useful updates}
\label{sec:structural_reference}
A binary single-hidden-layer ReLU margin can be written
\begin{equation}
 f_\theta(x)=c+\sum_{i=1}^{h}v_i\operatorname{ReLU}(w_i^\top x+b_i).
 \label{eq:relu_margin}
\end{equation}
On clean Boolean vertices, order-$k$ parity has a width-$k$ construction whose first-layer matrix has rank one: distinct thresholds along $\sum_j s_j$ reproduce the alternating labels. Interaction order and matrix rank therefore measure different properties. For zero hidden biases, $f(x)-f(-x)$ is linear~\cite{zhang2025}, precluding strictly correct classification of every vertex for odd parity $k\geq3$. These are architectural references, proved or specialized in Appendix~\ref{app:structure}; they do not establish the noisy task's Bayes attainability or a successful optimization path. All current synthetic comparisons retain trainable hidden biases.

Training restrictions can affect useful updates even when the restricted input is target-independent. With $a_i=w_{T,i}^\top x_T+w_{S,i}^\top x_S+w_{N,i}^\top x_N+b_i$ and $\delta=\partial\ell/\partial f$, one sample contributes
\begin{equation}
 \nabla_{w_{T,i}}\ell=\delta\,v_i\,\mathbf1\{a_i>0\}\,x_T.
 \label{eq:gradient_access}
\end{equation}
Nuisance can influence the activation gate and the prediction-dependent error $\delta$, and hence the updates on target coordinates. Optimizer state and matrix transformations add further coupling. This identity supplies concrete candidate pathways through which restricting nuisance adaptation could change relation learning. The experiments identify the outcome-level dependence, without selecting one of these pathways as its established mediator. Appendix~\ref{app:gate} gives a Gaussian gate-sensitivity reference and explains why parameter displacement alone does not measure functional importance.

\subsection{Selection, pairing, and reporting}
\label{sec:selection}
FWSTEP01 chooses a learning rate using validation performance on search seeds 11 and 12; seeds 21--23 provide independent reporting runs, each with a validation-selected checkpoint. FWSTEP02 repeats selection for each training condition over three neighboring or extrapolated rates around the earlier choice. Its comparisons use its own reselected full-training baselines. Within a seed, configurations share generated data, initialization, and batch order. Validation retains the training cue correlation; evaluation reduces that correlation in the three cue-bearing tasks. The nuisance-heavy task instead evaluates fresh examples from the same distribution.

FWSTEP03 uses seeds 42--44 for both learning-rate selection and reporting: it selects a common rate by the mean of each seed's best validation accuracy, then reports the corresponding independent-test predictions. It does not use separate search seeds. In all these studies, accuracy selects checkpoints and rates, with validation loss for ties; test scores are not selection criteria. Synthetic summaries use three-seed means and sample standard deviations, with gains computed before rounding. This is variation across seeds, not a claim of statistical significance.

The expanded optimizer comparison follows FWSTEP01's protocol. CIFAR-10 is a separate one-seed, fixed-learning-rate comparison with validation checkpoint selection. The retained earlier eight-class control instead fixes the learning rate and epoch across conditions. Section~\ref{sec:experiments} reports the budgets, selected outcomes, and search-boundary audits; Appendix~\ref{app:reproduction} records the grids and source mapping. These protocols make the tested optimization conditions part of the claim, while preserving the distinction between available information, representational capacity, acquired prediction, and its dependence on nuisance or cues.

\section{Experiments and Discussion}
\label{sec:experiments}

The experiments examine which target relations different optimizer configurations acquire and use, and how those outcomes change when adaptation to nuisance inputs is constrained. Composite tasks expose cue dependence, rare-mode performance, and both shared and optimizer-dependent responses to training interventions. A separate strict-parity sweep characterizes generalization under a fixed shallow architecture and training budget. The expanded optimizer comparison tests the persistence of stage-level failures across update configurations, while CIFAR-10 provides an external architecture comparison.

\subsection{Experimental scope and reporting protocol}
\label{sec:fw_protocol}

FWSTEP01/02 implement the four composite tasks in Section~\ref{sec:task_families} with a $3124\!\rightarrow\!256\!\rightarrow\!2$ ReLU MLP, 8,192 training, 2,048 validation, and 4,096 test examples per seed, batch size 128, and 100 epochs. The optimizer implementations follow Section~\ref{sec:update_interventions}; all synthetic runs use zero weight decay. SGDM includes runs labeled \texttt{sgd} in the source files.

FWSTEP01 searches five learning rates using seeds 11 and 12, then reports seeds 21--23. FWSTEP02 searches three neighboring or extrapolated rates separately for each intervention and uses its own reselected full-training baselines. Both use validation accuracy, with loss for ties, as specified in Section~\ref{sec:selection}. Thus the intervention comparisons concern the shared selection protocol, with condition-specific rates and checkpoints.

Unless otherwise stated, synthetic results are three-seed means and sample standard deviations. Differences are percentage points (pp), calculated from unrounded seed-level results. The seed is the experimental replicate. Stage labels retain the definition in Section~\ref{sec:task_families} and the conditional audit in Appendix~\ref{app:stage_audit}; strict interaction order is reserved for parity. The nuisance-heavy task evaluates in distribution, while the three cue-bearing tasks reduce cue correlation at test time.

Table~\ref{tab:fw_baseline} summarizes the initial composite comparison. The higher-order and mixed tasks have fourth-stage accuracy near chance for all three optimizers. Thus, their overall scores alone do not establish acquisition of the hardest component. These observations motivate direct intervention on the training conditions.

\begin{table}[H]
\centering\small
\caption{FWSTEP01 baseline test accuracy (\%, mean $\pm$ sample SD, seeds 21--23). The higher-order nuisance task is evaluated in distribution; the other tasks reduce spurious-cue correlation at test time.}
\label{tab:fw_baseline}
\begin{tabular}{lccc}\toprule
Task & SGDM & Adam & Muon\\\midrule
Distributed votes & $67.19 \pm 1.24$ & $67.47 \pm 0.91$ & $62.19 \pm 3.18$ \\
Higher-order + nuisance & $63.72 \pm 0.11$ & $62.35 \pm 0.79$ & $63.06 \pm 1.28$ \\
Rare modes & $85.86 \pm 0.34$ & $71.88 \pm 0.24$ & $86.81 \pm 0.49$ \\
Mixed evidence & $74.45 \pm 0.14$ & $73.05 \pm 0.86$ & $74.96 \pm 0.45$ \\

\bottomrule\end{tabular}
\end{table}
\FloatBarrier

\subsection{Nuisance-connected trainability changes rule generalization}
\label{sec:fw_intervention}

The higher-order nuisance task contains 80 target-signal coordinates, 320 coordinates encoding irrelevant relations, and 2,724 background-noise coordinates. Fourth-stage examples constitute half of the data. Let $N$ denote the 3,044 nuisance coordinates and $W$ the first-layer weight matrix. Following Section~\ref{sec:update_interventions}, the freezing intervention enforces
\begin{equation}
 W_{:,N}^{(t)}=W_{:,N}^{(0)}\qquad\text{throughout training}.
 \label{eq:fw_freeze}
\end{equation}
The selected gradients are zeroed, and the selected columns are restored after every optimizer step, including coupled matrix updates. Raw inputs, network architecture, initialization seed, and batch ordering are retained. Nuisance still contributes its initial random projection to hidden preactivations; the remaining parameters continue to adapt. This fixes $3{,}044\times256=779{,}264$ first-layer weights, reducing the number of trainable parameters from 800,514 to 21,250. The forward architecture is unchanged, while the space of permitted parameter updates is substantially reduced.

\begin{figure}[H]
\centering\includegraphics[width=0.96\linewidth]{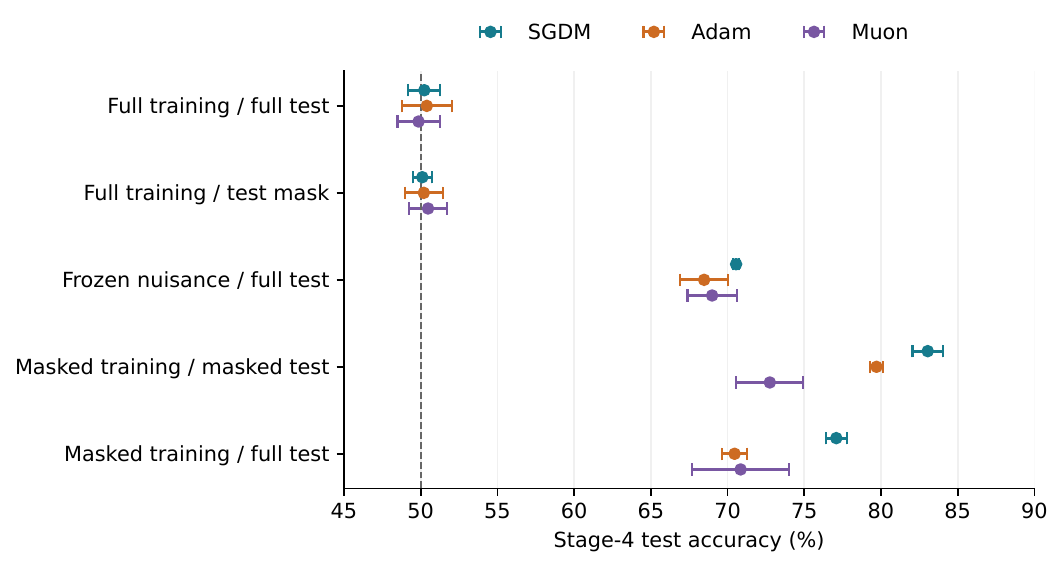}
\caption{\textbf{Constraining training recovers performance that test-time masking does not.} Points show fourth-stage test means; horizontal bars show sample SD across three seeds. The dashed line marks chance. All three methods improve when nuisance-connected weights are frozen despite continued exposure to nuisance.}
\label{fig:fw_core}
\end{figure}

\paragraph{Freezing during training and masking after training have different effects.}
Table~\ref{tab:fw_core} and Figure~\ref{fig:fw_core} show the central result. Full training gives fourth-stage accuracy of 50.24\%, 50.40\%, and 49.86\% for SGDM, Adam, and Muon. Masking nuisance only at evaluation leaves accuracy near 50\%. Freezing nuisance-connected weights throughout training instead yields 70.56\%, 68.47\%, and 69.00\%, corresponding to gains of 20.32, 18.08, and 19.13 pp. Each optimizer improves in every reporting seed. The network therefore obtains a substantially different generalizing function when adaptation to the same nuisance inputs is restricted.

\begin{table}[H]
\centering\small\setlength{\tabcolsep}{4pt}
\caption{Fourth-stage test accuracy in FWSTEP02 (\%, three-seed mean $\pm$ SD). All rows concern the same task. Frozen training retains full inputs during training, validation, and the reported test. Masked training zeros nuisance in training and validation; its two test evaluations use the same selected model. Each training condition has its own validation-selected rate and checkpoint.}
\label{tab:fw_core}
\begin{tabular}{lccc}\toprule
Training / evaluation condition & SGDM & Adam & Muon\\\midrule
Full training / full test & $50.24 \pm 1.04$ & $50.40 \pm 1.63$ & $49.86 \pm 1.37$ \\
Full training / test mask & $50.11 \pm 0.61$ & $50.20 \pm 1.22$ & $50.48 \pm 1.24$ \\
Frozen nuisance / full test & $70.56 \pm 0.20$ & $68.47 \pm 1.55$ & $69.00 \pm 1.61$ \\
Masked training / masked test & $83.05 \pm 0.99$ & $79.71 \pm 0.42$ & $72.76 \pm 2.18$ \\
Masked training / full test & $77.09 \pm 0.68$ & $70.46 \pm 0.84$ & $70.85 \pm 3.15$ \\

\bottomrule\end{tabular}
\end{table}

\paragraph{Removing exposure produces an additional effect.}
Zeroing nuisance throughout training and evaluation raises fourth-stage accuracy further, to 83.05\%, 79.71\%, and 72.76\%. Restoring full inputs only at test time reduces these scores to 77.09\%, 70.46\%, and 70.85\%, but they remain well above full-training baselines. Recovery therefore has a training component that survives renewed nuisance exposure, alongside sensitivity to the evaluation inputs. SGDM has the highest score in the masked condition; Adam and Muon also recover substantial performance.

\paragraph{The joint intervention matters.}
Freezing only relation-noise connections gives 50.06--50.40\% fourth-stage accuracy across optimizers; freezing only background connections gives 49.72--50.50\%. Freezing both gives 68.47--70.56\%. Thus, neither individual block accounts for recovery in this setting. The search trajectories also show the same qualitative separation across all three rates: the mean peak fourth-stage validation accuracy remains approximately 51--52\% under full training, reaches 64--71\% under joint freezing, and 72--83\% under joint masking. These trajectory diagnostics use validation data and show that the separation extends beyond one selected checkpoint.

\paragraph{A fixed-condition experiment provides complementary support.}
The earlier eight-class task compares the same learning rate, epoch, and seed across interventions (Table~\ref{tab:fw_old_control}). AdamW rises from 44.73\% to 95.07\% overall when independent-noise connections are frozen, whereas test-only zeroing gives 48.38\%. Accuracy on factor $C$, its third-order target component, rises from 50.07\% to 97.49\%. The 60-coordinate spurious block remains present and trainable. SGDM and Shampoo start near 97\% and show smaller gains. This single-seed, common-epoch contrast complements the new multi-seed comparisons in which each condition is separately tuned.

\begin{table}[H]
\centering\small
\caption{Earlier eight-class intervention at fixed LR 0.05, epoch 10, seed 42. Full, Test zero, and Frozen are overall accuracies (\%); the last two columns measure factor $C$. This experiment has its own generator and $3124\!\rightarrow\!64\!\rightarrow\!8$ architecture.}
\label{tab:fw_old_control}
\begin{tabular}{lrrrrr}\toprule
Optimizer & Full & Test zero & Frozen & $C$: full & $C$: frozen\\\midrule
AdamW & 44.73 & 48.38 & 95.07 & 50.07 & 97.49\\
SGDM & 96.86 & 97.79 & 98.22 & 98.01 & 99.10\\
Shampoo & 96.90 & 97.71 & 98.28 & 97.99 & 99.12\\\bottomrule
\end{tabular}
\end{table}

Together, these results identify nuisance-connected trainability as an effective control on learning outcomes. The fourth-stage latent-prefix audit is also informative: even the complete third prefix predicts labels only at chance within this group (Appendix~\ref{app:stage_audit}). However, the recovered models have not yet been tested with their designated fourth-order target block removed. The established result is recovery on this group; its exact internal computation remains to be identified.
\FloatBarrier

\subsection{Optimizer advantages depend on the interference conditions}
\label{sec:fw_conditional}

\paragraph{Distributed evidence does not guarantee rejection of a concentrated cue.}
The vote task combines 17 genuine first-order evidence sources with one stronger cue that matches the label approximately 90\% of the time in training and 50\% in testing. Although its evidence sources are first order, the thresholded majority label is not a pure degree-one Boolean function. Adam's baseline accuracy exceeds SGDM's by only 0.28 pp, with seed-wise differences of mixed signs. All three methods have a large gap between cue-matching and cue-conflicting examples (Table~\ref{tab:fw_cues}). Muon is most sensitive here: reflecting the spurious signal changes 74.14\% of predictions, and its conflict accuracy is 24.72\%.

Training with the cue removed raises all three methods to about 80\%. The gains are 13.22 pp for SGDM, 12.33 pp for Adam, and 17.73 pp for Muon. The true evidence is learnable under the same architecture, but unrestricted training retains substantial dependence on the concentrated shortcut. These results distinguish learning distributed evidence from reliably selecting it in the presence of a competing cue.

\begin{table}[H]
\centering\small\setlength{\tabcolsep}{4pt}
\caption{Vote task: cue dependence and recovery (\%). Overall accuracies show mean $\pm$ SD; the final three columns show seed means for the full-trained baseline. Match/conflict refers to agreement between the cue and true label. Flip is the fraction of predictions changed by reflecting the cue. Cue removal is a training-time input mask in FWSTEP02; reflection diagnostics come from the matching FWSTEP01 baselines.}
\label{tab:fw_cues}
\begin{tabular}{lccrrr}\toprule
Optimizer & Baseline & Cue removed & Match & Conflict & Flip\\\midrule
SGDM & $67.19 \pm 1.24$ & $80.41 \pm 0.98$ & 98.16 & 35.68 & 62.26 \\
Adam & $67.47 \pm 0.91$ & $79.80 \pm 0.75$ & 96.21 & 38.25 & 58.24 \\
Muon & $62.19 \pm 3.18$ & $79.92 \pm 0.52$ & 99.03 & 24.72 & 74.14 \\

\bottomrule\end{tabular}
\end{table}

\paragraph{Cue reliance differs across target structures.}
The second-order mode task exhibits a different pattern (Table~\ref{tab:fw_second_cues}). Muon's cue-conflict accuracy is 83.44\%, and reflecting the cue changes 7.17\% of predictions, compared with 24.72\% and 74.14\% in the vote task. SGDM and Adam also retain substantial conflict accuracy in the second-order task. These diagnostics cover all test modes, rather than only the rarest mode. They show that pronounced cue dependence in one task does not define a fixed optimizer trait. The tasks also differ in training cue-label agreement (approximately 65\% here versus 90\% in the vote task), signal organization, and nuisance structure, so this contrast does not isolate minimum useful interaction order as its sole cause.

\begin{table}[H]
\centering\small
\caption{Second-order mode task: cue dependence of the full-trained FWSTEP02 baselines (\%, mean $\pm$ sample SD, seeds 21--23), evaluated across all modes. Match and conflict refer to cue-label agreement; Flip measures prediction changes under cue reflection. Adam uses the FWSTEP02 reselected baseline LR 0.0125. These are cue diagnostics, distinct from the rare-mode slice in Table~\ref{tab:fw_rare}.}
\label{tab:fw_second_cues}
\begin{tabular}{lrrr}\toprule
Optimizer & Match & Conflict & Flip\\\midrule
SGDM & $90.00 \pm 0.35$ & $81.69 \pm 1.00$ & $8.50 \pm 0.62$\\
Adam & $81.65 \pm 0.90$ & $72.26 \pm 1.62$ & $10.25 \pm 0.14$\\
Muon & $90.16 \pm 0.49$ & $83.44 \pm 1.46$ & $7.17 \pm 1.03$\\
\bottomrule\end{tabular}

\end{table}

\paragraph{The rare-mode advantage changes under background freezing.}
The three second-order modes occur in a $60/30/10$ ratio. In the FWSTEP02 baseline, Muon achieves 67.56\% on the rarest mode, compared with 55.58\% for SGDM and 55.96\% for Adam. Freezing background connections raises these accuracies to 84.37\%, 84.87\%, and 88.59\%, respectively (Table~\ref{tab:fw_rare}). Mode frequency and the target relation are unchanged, yet Muon no longer leads on average. Removing all nuisance during training also brings all three methods above 87\% on the rare mode.

\begin{table}[H]
\centering\small
\caption{Rarest-mode test accuracy in FWSTEP02 (\%, mean $\pm$ SD). All-nuisance removal masks relation noise and background while retaining the spurious cue. Adam's reselected baseline LR is 0.0125, versus 0.005 in FWSTEP01; these intervention comparisons use the FWSTEP02 baseline throughout.}
\label{tab:fw_rare}
\begin{tabular}{lccc}\toprule
Optimizer & Full baseline & Freeze background & Remove all nuisance\\\midrule
SGDM & $55.58 \pm 1.89$ & $84.87 \pm 2.40$ & $89.64 \pm 0.39$ \\
Adam & $55.96 \pm 0.16$ & $88.59 \pm 1.02$ & $90.06 \pm 1.63$ \\
Muon & $67.56 \pm 2.06$ & $84.37 \pm 1.96$ & $87.13 \pm 1.84$ \\

\bottomrule\end{tabular}
\end{table}

For the gap contrast in Eq.~\eqref{eq:optimizer_interaction}, background freezing changes the Muon--SGDM rare-mode gap from $+11.99$ to $-0.49$ pp, a reduction of 12.48 pp. The paired reductions are 12.99, 12.28, and 12.17 pp for seeds 21, 22, and 23. The frozen gaps are $-1.96$, $0.00$, and $+0.48$ pp: the original advantage disappears, without a consistent seed-wise reversal in favor of SGDM. Thus, background trainability changes the optimizer comparison within the same task, beyond improving each optimizer's score separately.

Accuracy also leaves an important aspect of this comparison unresolved. The FWSTEP01 accuracy-selected Muon model has mean test cross-entropy 47.97, compared with 0.627 for SGDM and 1.641 for Adam. Its higher rare-mode accuracy coexists with highly overconfident errors. The rare-mode advantage is consequently an accuracy result, not a general improvement in predictive probability quality.

\paragraph{A mixed task establishes a recovery boundary.}
Removing every non-target input block in the mixed task raises overall accuracy, but fourth-stage accuracy remains 52.41\%, 51.14\%, and 50.68\% for SGDM, Adam, and Muon (Table~\ref{tab:fw_boundary}). The strongest successful intervention from the previous task therefore does not explain every failure. This task also has a lower fourth-stage frequency (25\% versus 50\%) and weaker target signal relative to observation noise. Those factors vary together across tasks, so their separate contributions are not identified by this comparison. The result establishes a concrete boundary: suppressing the labeled interference is insufficient under this mixed task's remaining conditions.

\begin{table}[H]
\centering\small\setlength{\tabcolsep}{4pt}
\caption{Mixed-task recovery boundary (\%, mean $\pm$ SD). ``Removed'' zeros all non-target blocks during training, validation, and testing. Improvements in the overall score coexist with near-chance fourth-stage performance.}
\label{tab:fw_boundary}
\begin{tabular}{lcccc}\toprule
& \multicolumn{2}{c}{Overall accuracy} & \multicolumn{2}{c}{Stage-4 accuracy}\\
Optimizer & Full & Removed & Full & Removed\\\midrule
SGDM & $74.45 \pm 0.14$ & $77.95 \pm 0.68$ & $50.20 \pm 0.10$ & $52.41 \pm 1.30$ \\
Adam & $73.05 \pm 0.86$ & $77.45 \pm 0.46$ & $50.98 \pm 0.52$ & $51.14 \pm 0.37$ \\
Muon & $74.96 \pm 0.45$ & $77.01 \pm 0.54$ & $50.46 \pm 1.13$ & $50.68 \pm 1.50$ \\

\bottomrule\end{tabular}
\end{table}

\begin{figure}[H]
\centering\includegraphics[width=\linewidth]{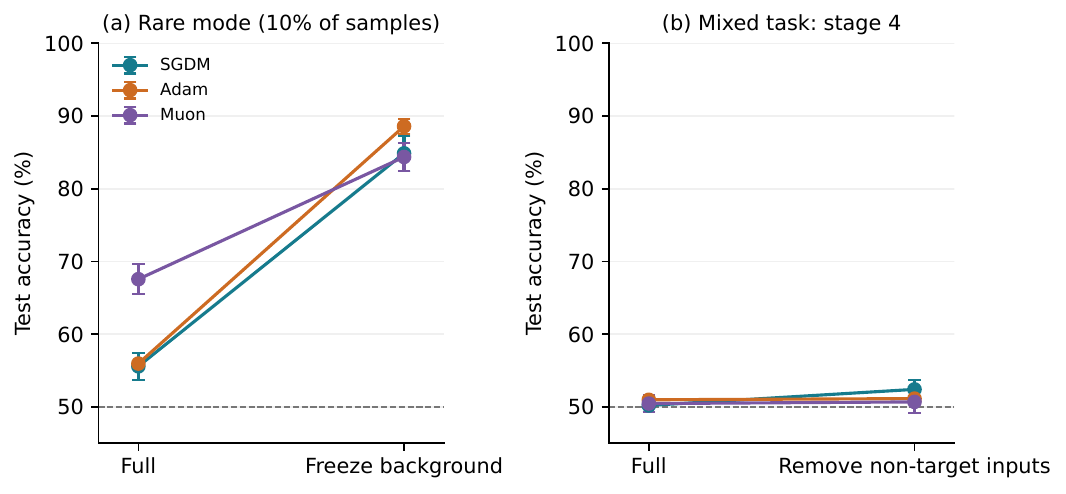}
\caption{\textbf{Conditional advantage and a recovery boundary.} (a) Background freezing changes the rare-mode ordering within the same task. (b) Removing all non-target inputs does not substantially recover the mixed task's fourth-stage group. Bars are sample SD across three seeds. The panels concern different tasks and different interventions.}
\label{fig:fw_conditional}
\end{figure}
\FloatBarrier

\subsection{Strict parity separates available information from achieved learning}
\label{sec:fw_strict}

FWSTEP03 removes spurious cues and independent nuisance coordinates and uses a strict target $y=\prod_{j=1}^{k}s_j$ with independent balanced latent signs. Every proper subset of these factors is uninformative about $y$. The inputs retain Gaussian observation noise with $\sigma=0.35$; ``signal only'' does not mean noiseless observations. The total clean signal norm is four, giving factor amplitude $4/\sqrt{k}$. Thus, increasing order also reduces the per-factor signal-to-noise ratio, which the Bayes reference explicitly tracks.

The architecture is fixed at $32\!\rightarrow\!64\!\rightarrow\!2$ (2,242 parameters), with 8,192 examples in each split and a 200-epoch budget. Seeds 42--44 determine data and initialization; optimizers receive the same batches within each seed. At each order and optimizer, a common learning rate is selected by mean best validation accuracy over these three seeds, after checkpoint selection within each run. These same seeds supply the independent-test results; unlike FWSTEP01/02, there is no separate set of learning-rate search seeds. Validation is recorded at epochs 1, 2, 5, and every ten epochs thereafter. The final grids contain seven SGDM, six Adam, and four Muon rates, including the supplemental low-rate SGDM search.

\begin{figure}[H]
\centering\includegraphics[width=\linewidth]{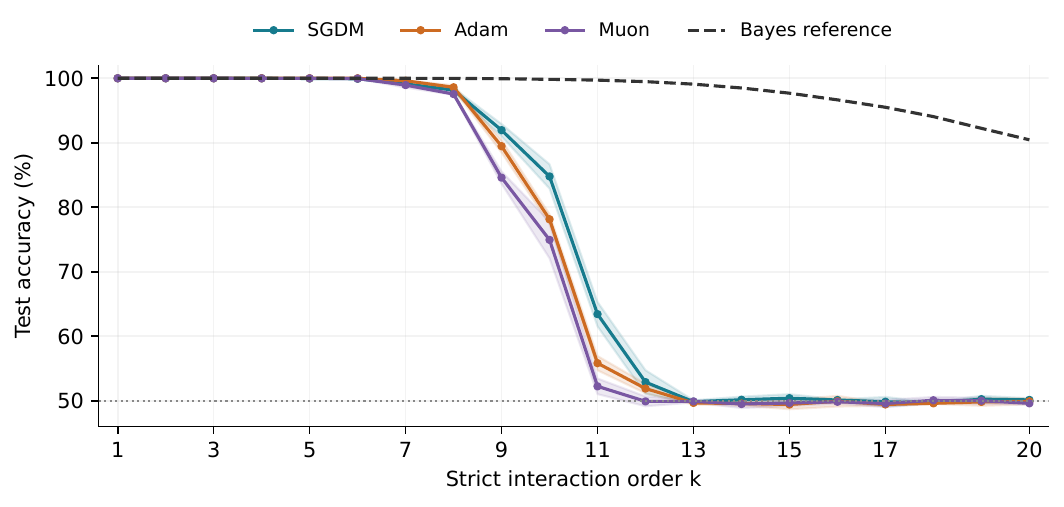}
\caption{\textbf{Strict parity learning from one to twenty factors.} Lines show three-seed test means at validation-selected rates; shading is $\pm$ one sample SD. The Bayes reference averages the larger class posterior on test observations using the known generator. It measures available predictive information, not performance proven attainable by the fixed MLP.}
\label{fig:fw_strict}
\end{figure}

All methods exceed 99.9\% mean test accuracy through order six (Appendix~\ref{app:reproduction}). Adam is slightly ahead at orders seven and eight. At orders nine through eleven, SGDM leads Adam in each reporting seed, with mean differences of 2.48, 6.64, and 7.60 pp (Table~\ref{tab:fw_parity}). At order ten, the test scores are 84.78\%, 78.14\%, and 74.96\%; at order eleven they are 63.44\%, 55.84\%, and 52.25\%. These differences locate a transition region within the tested shallow architecture and budget.

\begin{table}[H]
\centering\small
\caption{Strict parity transition (\%, mean $\pm$ sample SD, seeds 42--44). The Bayes column is the mean posterior-based reference. All twenty orders and their selected rates are retained in the accompanying data.}
\label{tab:fw_parity}
\begin{tabular}{rcccr}\toprule
Order & SGDM & Adam & Muon & Bayes\\\midrule
7 & $99.13 \pm 0.08$ & $99.61 \pm 0.08$ & $98.94 \pm 0.40$ & 100.00 \\
8 & $98.16 \pm 0.30$ & $98.60 \pm 0.32$ & $97.55 \pm 0.11$ & 99.98 \\
9 & $91.95 \pm 0.96$ & $89.47 \pm 0.97$ & $84.60 \pm 0.96$ & 99.93 \\
10 & $84.78 \pm 1.88$ & $78.14 \pm 0.79$ & $74.96 \pm 2.86$ & 99.84 \\
11 & $63.44 \pm 1.90$ & $55.84 \pm 1.11$ & $52.25 \pm 1.21$ & 99.71 \\
12 & $52.90 \pm 1.85$ & $51.90 \pm 0.74$ & $49.92 \pm 0.76$ & 99.48 \\

\bottomrule\end{tabular}
\end{table}

The difference is not captured by training fit alone. At the selected order-eleven checkpoints, mean online training accuracies are 78.59\%, 79.56\%, and 79.63\% for SGDM, Adam, and Muon, despite the larger separation in test accuracy. These online values aggregate predictions during the selected epoch rather than reevaluating a fixed checkpoint; they establish a descriptive fit--generalization separation without claiming complete memorization.

At orders thirteen through twenty, all three methods remain within 0.57 pp of chance, while the Bayes reference decreases from approximately 99.09\% to 90.46\%. Observations therefore retain considerable target information even where the evaluated learners fail to use it. This gap does not by itself separate insufficient representation, insufficient training time, and unfavorable update dynamics.

The learning-rate and budget audits sharpen this interpretation. At order ten, extending SGDM's search changes the selected rate from 0.05 to 0.02 and improves mean best validation accuracy from 81.45\% to 85.36\%. Both SGDM and Adam select rates internal to their final grids at orders nine through eleven. Muon instead selects its lowest tested rate, 0.005, at every order from seven to twelve. Its lower-rate regime remains untested, so the observed Muon deficit cannot be interpreted as a comparison after exhaustive tuning. Extending its grid downward in this transition region, while retaining the same seeds, budget, and validation criterion, is the direct test of whether its relative position persists. Adam also selects its upper boundary, 0.02, at order seven and its lower boundary, 0.0005, at order eight; these boundary cases accompany the reported crossover.

SGDM's order-eleven checkpoints occur at epochs 180--200, so additional training is also a relevant open direction. Of 1,020 runs, 1,019 finish normally; the one nonfinite-loss run, at order six with SGDM LR 0.2 and seed 44, is not part of that order's selected LR 0.05 results. The learning map describes attained performance under the stated searches and finite budget.
\FloatBarrier

\subsection{Expanding the optimizer set leaves the fourth-stage bottleneck}
\label{sec:fw_expanded}

The FWSTEP01 extension adds 17 configurations to the three representative methods, yielding twenty configurations and 240 seed-level reports over four tasks. It includes momentum variants, adaptive methods, matrix preconditioners, and averaging variants. The new methods use five-rate grids, two search seeds, and the same three reporting seeds, task generators, architecture, and 100-epoch budget. The original three methods are reused from FWSTEP01 rather than counted as new runs.

The extension improves some overall scores without recovering the fourth-stage group. In the higher-order nuisance task, Nesterov reaches 64.44\% overall versus SGDM's 63.72\%, but all twenty selected models have fourth-stage means between 49.86\% and 50.53\%. In the mixed task, the corresponding range is 49.74--51.30\%. By comparison, the training-permission intervention raises the higher-order nuisance task to 68.47--70.56\% while retaining full inputs. Within the tested configurations, changing nuisance adaptation alters this failure far more than expanding the unrestricted optimizer list.

Other differences remain task-specific. RAdam and ADOPT reach 71.44\% and 71.31\% in the vote task, above Adam's 67.47\%, but their cue-conflict accuracies remain 46.85\% and 47.26\%. In the rare-mode task, Muon-EMA and Muon reach 68.58\% and 67.56\% on the rarest mode; the highest value outside these two configurations is 55.80\%. Thus, the expanded comparison preserves important subgroup differences while challenging a fixed assignment of one optimizer to each task type.

The coverage is defined by the search protocol: 41 of 68 added task--configuration combinations select a grid boundary, including an upper-boundary choice for 14 of 17 new configurations on the rare-mode task. The extension does not retain stage-specific trajectories for all methods, so its fourth-stage statement concerns selected checkpoints. AdamW duplicates Adam exactly in the zero-weight-decay runs and is an implementation check rather than independent performance evidence. Full numerical results and selected configurations are retained with the manuscript data and summarized in Appendix~\ref{app:reproduction}.
\FloatBarrier

\subsection{CIFAR-10 provides an external architecture comparison}
\label{sec:fw_cifar}

We also examine nine update configurations on Shallow-3 and VGG13-BN. The former has three convolutional stages and 227,370 parameters; the latter has ten convolutional layers and 9,675,722 parameters. Both use a fixed 45,000/5,000 training/validation split of CIFAR-10 and its 10,000-image test set, seed 42, 200 epochs, batch size 128, cosine learning-rate decay, weight decay $5\times10^{-4}$, random cropping, and horizontal flipping. Each configuration's checkpoint is selected by validation accuracy. Here SGDM uses Nesterov momentum 0.9; learning rates are fixed rather than searched per architecture.

\begin{table}[H]
\centering\small
\caption{CIFAR-10 test accuracy at each configuration's validation-selected checkpoint (\%, seed 42). A slash denotes convolution / classifier-head update rules. PAdam's negative-$p$ variants are the experimental implementations used in these runs; Muon uses an AdamW fallback on vector parameters.}
\label{tab:fw_cifar}
\begin{tabular}{lrr}\toprule
Configuration & Shallow-3 & VGG13-BN\\\midrule
SGDM & 88.02 & 93.19 \\
Adam & 87.42 & 92.08 \\
Muon & 86.72 & 92.75 \\
PAdam $(p=0.25)$ & 87.75 & 93.15 \\
PAdam $(p=-0.25)$ & 87.19 & 91.47 \\
PAdam $(p=-0.50)$ & 87.25 & 91.92 \\
Muon / SGDM & 86.04 & 93.47 \\
Muon / PAdam $(-0.25)$ & 86.98 & 93.33 \\
Muon / PAdam $(-0.50)$ & 87.88 & 93.47 \\

\bottomrule\end{tabular}
\end{table}

The highest observed shallow-network test score is 88.02\% with SGDM. In VGG13-BN, convolutional Muon with a SGDM head or a PAdam $p=-0.50$ head reaches 93.47\%, versus 93.19\% for SGDM. The direction of the head intervention differs across architectures: replacing Muon's matrix update in the classifier by SGDM changes test accuracy from 86.72\% to 86.04\% in Shallow-3, and from 92.75\% to 93.47\% in VGG13-BN. The classifier's parameter share also changes from 58.9\% to 2.77\%, together with network depth and size. This comparison establishes an observed architecture dependence without isolating which architectural change produces it.

Learning curves provide a further descriptive parallel. At epoch ten, Muon's validation accuracy exceeds SGDM's in both networks (80.84\% versus 69.22\% in Shallow-3; 89.22\% versus 74.38\% in VGG13-BN). By epoch 200, SGDM is ahead in both comparisons. Early progress therefore need not predict the eventual ordering. Epoch counts do not imply equal wall-clock cost. With one seed, no learning-rate sweep, and no annotated interaction orders or nuisance interventions, these image results serve as external performance evidence; the controlled synthetic experiments support the claims about relation acquisition.
\FloatBarrier

\subsection{Findings on optimizer-dependent rule acquisition}
\label{sec:fw_discussion}

\paragraph{Shared failures and relative advantages depend on trainable pathways.}
Holding nuisance-connected weights at initialization recovers fourth-stage performance for all three representative optimizers, even though nuisance remains in the inputs and hidden preactivations. Evaluation-time masking does not reproduce this recovery. Within the rare-mode task, restricting background adaptation also removes Muon's original advantage over SGDM, reducing their gap by 12.48 pp. The target relation and mode frequency are unchanged. Together with the earlier fixed-condition contrast, these results identify training permissions as a control on both rule generalization and optimizer comparisons.

\paragraph{Formation, selection, and expression require separate diagnosis.}
The vote and second-order tasks show markedly different cue reliance, while the mixed task retains a failure after all designated non-target inputs are removed. Strict parity exhibits optimizer separation without spurious cues or extraneous nuisance coordinates, followed by a shared loss of generalization despite substantial available information. A useful account must therefore explain relation- and mode-specific performance, successful recovery, persistent failure, and changing optimizer gaps. Test-time masking alone cannot distinguish absent internal features from useful features that are poorly gated or read out. Reduced cue sensitivity, by itself, does not identify a higher-order self-review mechanism. The large reduction in trainable parameters under freezing also motivates controls that match their number, to separate the role of connection location from the amount of permitted adaptation.

\paragraph{The next mechanistic target is adaptation that can be identified in advance.}
The present interventions use generator knowledge to select input blocks. A practical extension must replace that knowledge with training-observable criteria. A decisive test would branch identical intermediate checkpoints, constrain updates using a proposed criterion, and compare subsequent relation acquisition under matched data and compute while checking preservation of useful behavior. The observed recoveries, optimizer reorderings, and persistent failures provide positive and negative controls for testing such prospective interventions.

\section{Conclusion}
\label{sec:conclusion}

Optimizer-dependent rule acquisition is jointly shaped by the target evidence, competing inputs, and permitted adaptation. Across controlled shallow-network tasks, the same nominal optimizer advantage can be large on a rare mode, modest in the aggregate, and absent after a change in background trainability. Conversely, twenty unrestricted configurations can share a difficult-stage failure that is substantially recovered by restricting nuisance-connected adaptation. These findings turn optimizer comparison into a study of which training conditions sustain a useful predictive function.

The central distinction is between exposure and adaptation. Keeping nuisance inputs present while fixing their first-layer connections recovers behavior that removing the inputs only after training does not. The strict-parity sweep supplies a complementary comparison: without spurious cues or extra nuisance coordinates, optimizers still separate at intermediate orders before reaching a shared finite-budget failure. The mixed task shows that removal of designated interference is not sufficient in every setting. A mechanism of rule acquisition must explain all three observations: conditional recovery, changing optimizer gaps, and persistent failure.

The next step is to predict useful interventions from training-observable signals. Activation margins, update alignment, optimizer state, and readout provide specific candidates, while the present positive and negative controls test whether an explanation goes beyond a correlation with final accuracy. Matched-parameter-count controls, necessity tests on recovered target inputs, and the lower-rate Muon extension would sharpen the corresponding claims. A successful practical method must identify which adaptations support reliable relations without generator-provided groups, and preserve useful behavior when those adaptations are constrained or reopened. The contribution established here is the experimentally demonstrated dependence that such a method must account for: the rules acquired by a fixed architecture change with both its update rule and the permissions given to that rule.

\clearpage
\appendix
\section{Statistical and structural references}
\label{app:structure}

\subsection{Parity posterior and Bayes accuracy}
\label{app:bayes}
In Eq.~\eqref{eq:strict_parity}, the Gaussian likelihood ratio for one factor is $\exp(2\alpha_k u_j^\top x_j/\sigma^2)$. With equal sign priors,
\[
 \mathbb E[s_j\mid x_j]=\tanh(\alpha_k u_j^\top x_j/\sigma^2).
\]
Disjoint independent channels make the posterior over factors factorize, giving $\mathbb E[y\mid x]=\prod_j\mathbb E[s_j\mid x_j]$ and the posterior in Eq.~\eqref{eq:parity_bayes}. Each sign decision has error probability $q_k=\Phi(-\alpha_k/\sigma)$. Parity is correct exactly when an even number of independent sign errors occurs, so $A_k^\star=[1+(1-2q_k)^k]/2$. This is a population reference. The reported empirical reference averages posterior confidence on the actual test observations; its sampling fluctuations can differ from those of the oracle's hard-label accuracy.

The generator balances labels in each finite split and then draws the free signs, enforcing the last sign to match parity. Its single-example population law is the independent balanced-factor law above. Train, validation, and test observations are independent draws within the stated balancing scheme. A latent combination can recur across splits, especially at low order; held-out performance here concerns fresh noisy observations, not a split that excludes all previously seen latent combinations.

\subsection{Clean parity with rank-one first-layer weights}
For $s\in\{-1,+1\}^k$, write $t=\sum_j s_j$, $t_\ell=-k+2\ell$, and $p_\ell=(-1)^{k-\ell}$. At $t=t_\ell$ there are $\ell$ positive factors, hence parity is $p_\ell$. Let $d_\ell=(p_{\ell+1}-p_\ell)/2$ and define
\begin{equation}
 h(t)=p_0+d_0\operatorname{ReLU}(t+k)
 +\sum_{\ell=1}^{k-1}(d_\ell-d_{\ell-1})\operatorname{ReLU}(t-t_\ell).
 \label{eq:clean_construction}
\end{equation}
The slope changes telescope, so the slope on $[t_\ell,t_{\ell+1}]$ is $d_\ell$. Starting from $h(-k)=p_0$, induction gives $h(t_\ell)=p_\ell$. All $k$ hidden units use the same projection direction with different biases, so the first-layer matrix has rank one. For clean observed blocks, substitute $t=\sum_j u_j^\top x_j/\alpha_k$. This proves clean-vertex representability within width 64 for all scanned orders, without identifying a successful training trajectory. With observation noise, this projection has noise standard deviation $\sigma\sqrt{k}/\alpha_k$; the construction does not show Bayes-optimal prediction on those noisy observations.

\subsection{The odd-parity restriction without hidden bias}
The established linear odd-component property~\cite{zhang2025} follows from $\operatorname{ReLU}(a)-\operatorname{ReLU}(-a)=a$. With $b_i=0$ in Eq.~\eqref{eq:relu_margin},
\[
 f(x)-f(-x)=\sum_i v_iw_i^\top x.
\]
For odd $k\geq3$, suppose $y(s)f(s)>0$ at every Boolean vertex. Since $y(-s)=-y(s)$, correct classification of each antipodal pair would imply $y(s)a^\top s>0$ everywhere for a vector $a$. But
\[
 \mathbb E[y(s)a^\top s]=\sum_j a_j\mathbb E\!\left[\prod_{i\neq j}s_i\right]=0,
\]
contradicting strict positivity. Output bias cancels in the difference. Even parity is invariant under global reversal and is not excluded by this argument. The reference concerns an architectural restriction; it is not an explanation of the remaining differences among optimizers with trainable hidden biases.

\subsection{Independent background, displacement, and gates}
\label{app:gate}
Conditioning on signed labels, independent zero-mean background makes cross-example terms vanish in $\mathbb E\|n^{-1}\sum_i y_i b_i\|^2$, leaving $n$ contributions of $d_B\sigma_B^2/n^2$. This proves Eq.~\eqref{eq:empirical_noise}. It applies to the Gaussian background, not automatically to arbitrary structured distractors.

A descriptive adaptation statistic retained in the earlier experiment is
\[
 D_N(t)=\frac{\|W_{:,N}(t)-W_{:,N}(0)\|_F^2}{\|W(t)-W(0)\|_F^2},
\]
when the denominator is nonzero. It measures allocation of squared displacement, not information, memory capacity, or functional importance. For a fixed trained unit on a fresh test observation, let $m$ be its preactivation excluding Gaussian background and $\xi=w_B^\top x_B$. Conditional on the remaining inputs, $\xi\sim\mathcal N(0,s_B^2)$ with $s_B=\sigma_B\|w_B\|_2$. For $s_B>0$,
\begin{equation}
 \Pr[\mathbf1\{m+\xi>0\}\neq\mathbf1\{m>0\}\mid m]
 =\Phi(-|m|/s_B).
 \label{eq:gate_flip}
\end{equation}
For $m>0$, a flip requires $\xi\leq-m$; for $m<0$, it requires $\xi>-m$. At $m=0$ the probability is one half under the inactive-at-zero convention; with $s_B=0$ it is zero. This fresh-input statement cannot be applied unchanged to the same training noise on which weights were fitted. A gate can also remain open while transmitting background variation. The expression motivates activation diagnostics rather than establishing the mediator of the reported recoveries.

\section{Composite-stage information audit}
\label{app:stage_audit}
For $K=4$, the staged score is
\[
 H=4(q_{1,1}+q_{1,2}+q_{1,3})+2(q_{2,1}+q_{2,2}+q_{2,3})
 +(q_{3,1}+q_{3,2}+q_{3,3})+1.5q_{4,1}.
\]
The ten signs $q$ are independent and balanced before stage reweighting. The atom mapping $z_{1,c}=q_{1,c}$ and $z_{j,c}=q_{j,c}q_{j-1,c}$ is invertible along each chain. Enumeration of all $2^{10}$ configurations gives the conditional optimal accuracies in Table~\ref{tab:prefix_audit}. Reweighting stages leaves these within-stage quantities unchanged.

\begin{table}[H]
\centering\small
\caption{Exact latent-prefix audit of the staged generator. Columns give the best possible conditional accuracy (\%) from atoms through prefix $p$, with stage $J$ known to the diagnostic. No observation noise or fitted model is involved. Stage is not an input supplied to the trained network.}
\label{tab:prefix_audit}
\begin{tabular}{lrrrrrr}\toprule
Stage $J$ & Configurations & $p=0$ & $p=1$ & $p=2$ & $p=3$ & $p=4$\\\midrule
1 & 256 & 50 & 100 & 100 & 100 & 100\\
2 & 384 & 50 & 100 & 100 & 100 & 100\\
3 & 192 & 50 & 75 & 100 & 100 & 100\\
4 & 192 & 50 & 50 & 50 & 50 & 100\\\bottomrule
\end{tabular}
\end{table}

The stage certificate asks whether a prefix is sufficient before later evidence is revealed. Conditioning on the event that a particular stage is first certifiable supplies additional information about the later terms. The two questions need not have the same answer. Accordingly, a stage label cannot be read as a proven minimum degree for a learned predictor. The fourth-stage balance is a useful task check, but determining whether a recovered model needs its designated fourth-order atom still requires a model-level necessity intervention.

\section{Reproducibility and complete selected results}
\label{app:reproduction}
\subsection{Composite configurations}
The descriptive task names map to \texttt{adam\_low\_order}, \texttt{sgd\_high\_order}, \texttt{muon\_rare\_modes}, and \texttt{mixed\_review}, respectively. Names in the source are preset identifiers, not conclusions about optimizer superiority. All use view dimension eight, input dimension 3,124, and code seed 20261001.

\begin{table}[H]
\centering\small
\caption{Input composition and observation parameters. Dimensions are listed as target/spurious/relational nuisance/background; $(a,\sigma)$ specifies a factor channel. A dash denotes an absent block.}
\label{tab:task_parameters}
\begin{tabular}{lrrrr}\toprule
Setting & Votes & Nuisance & Rare modes & Mixed\\\midrule
Dimensions & 136/8/0/2980 & 80/0/320/2724 & 48/8/48/3020 & 80/80/160/2804\\
True $(a,\sigma)$ & (1.15,.85) & (1.65,.60) & (1.55,.65) & (1.50,.80)\\
Spurious $(a,\sigma)$ & (1.80,.55) & --- & (1.00,1.00) & (1.50,.80)\\
Relational $(a,\sigma)$ & --- & (1.50,.80) & (1.30,.80) & (1.50,.80)\\
Background $\sigma_B$ & .15 & .55 & .35 & .35\\
Cue match, train/test & .90/.50 & --- & .65/.50 & .78/.50\\\bottomrule
\end{tabular}
\end{table}

The nuisance task has four independent distractor relations at each constituent order 1--4 and stage proportions $(.10,.15,.25,.50)$. The mixed task has two distractors and one spurious relation per order, with equal stage proportions. Its order-$j$ spurious relation has the stated training agreement only on stage $j$; elsewhere agreement is one half. The mode task has two distractors at each order 1--2. Spurious and distractor products are encoded through separately observed factors. Validation cue agreement matches training. In the staged generator, cue-match counts are rounded to the nearest requested count, and stage counts are allocated deterministically before within-stage sampling and shuffling. Vote and mode tasks draw cue agreement independently with the specified probability.

\subsection{Search grids and selected checkpoints}
FWSTEP01 uses
\begin{align*}
 \Lambda_{\rm SGDM}&=\{.005,.01,.02,.05,.1\},\\
 \Lambda_{\rm Adam}&=\{.0002,.0005,.001,.002,.005\},\\
 \Lambda_{\rm Muon}&=\{.001,.002,.005,.01,.02\}.
\end{align*}
FWSTEP02 retains the earlier selected rate and its two neighbors, geometrically extrapolating a missing neighbor at an endpoint. Each intervention selects anew using seeds 11 and 12, then runs seeds 21--23. The archive retains the complete plan, including each task--optimizer--intervention grid and the validation-selected reporting epochs.

The final strict-parity grids, including the SGDM supplement, are
\begin{align*}
 \Lambda_{\rm SGDM}&=\{.005,.01,.02,.05,.1,.2,.5\},\\
 \Lambda_{\rm Adam}&=\{.0005,.001,.002,.005,.01,.02\},\\
 \Lambda_{\rm Muon}&=\{.005,.01,.02,.05\}.
\end{align*}
Evaluation is recorded at epochs 1, 2, 5, and each multiple of ten through 200. The data archive contains all selected rates, seed-level checkpoints, and boundary flags. Muon selects the lower endpoint at orders 7--12; the untested lower-rate regime is a specific outstanding coverage gap.

\subsection{All twenty strict orders}
\begin{table}[H]
\centering\small
\caption{Complete FWSTEP03 selected test accuracies (\%, mean $\pm$ sample SD, three seeds). The final column is the posterior-based Bayes reference. Learning-rate selection uses validation, as specified in Section~\ref{sec:selection}.}
\label{tab:all_parity}
\begin{tabular}{rrrrr}\toprule
Order & SGDM & Adam & Muon & Bayes\\\midrule
1 & $100.00\pm0.00$ & $100.00\pm0.00$ & $100.00\pm0.00$ & 100.00\\
2 & $100.00\pm0.00$ & $100.00\pm0.00$ & $100.00\pm0.00$ & 100.00\\
3 & $100.00\pm0.00$ & $100.00\pm0.00$ & $100.00\pm0.01$ & 100.00\\
4 & $100.00\pm0.00$ & $100.00\pm0.00$ & $99.99\pm0.00$ & 100.00\\
5 & $99.97\pm0.01$ & $100.00\pm0.00$ & $99.99\pm0.02$ & 100.00\\
6 & $99.93\pm0.06$ & $100.00\pm0.01$ & $99.95\pm0.02$ & 100.00\\
7 & $99.13\pm0.08$ & $99.61\pm0.08$ & $98.94\pm0.40$ & 100.00\\
8 & $98.16\pm0.30$ & $98.60\pm0.32$ & $97.55\pm0.11$ & 99.98\\
9 & $91.95\pm0.96$ & $89.47\pm0.97$ & $84.60\pm0.96$ & 99.93\\
10 & $84.78\pm1.88$ & $78.14\pm0.79$ & $74.96\pm2.86$ & 99.84\\
11 & $63.44\pm1.90$ & $55.84\pm1.11$ & $52.25\pm1.21$ & 99.71\\
12 & $52.90\pm1.85$ & $51.90\pm0.74$ & $49.92\pm0.76$ & 99.48\\
13 & $49.86\pm0.23$ & $49.68\pm0.26$ & $49.91\pm0.18$ & 99.09\\
14 & $50.17\pm0.49$ & $49.58\pm0.32$ & $49.51\pm0.63$ & 98.49\\
15 & $50.41\pm0.63$ & $49.43\pm0.73$ & $49.67\pm0.47$ & 97.69\\
16 & $50.13\pm0.10$ & $49.94\pm0.83$ & $49.85\pm0.28$ & 96.65\\
17 & $49.86\pm0.75$ & $49.44\pm0.22$ & $49.58\pm0.56$ & 95.50\\
18 & $49.69\pm0.05$ & $49.61\pm0.09$ & $50.09\pm0.52$ & 94.06\\
19 & $50.26\pm0.53$ & $49.84\pm0.62$ & $50.02\pm0.47$ & 92.24\\
20 & $50.15\pm0.25$ & $49.92\pm0.46$ & $49.61\pm0.22$ & 90.46\\
\bottomrule
\end{tabular}

\end{table}

\subsection{The expanded optimizer comparison}
\begin{table}[H]
\centering\scriptsize
\caption{All twenty configurations at their FWSTEP01 or appendix-selected checkpoints (three-seed test means, \%). ``Nuisance'' and ``Mixed'' report overall accuracy; ``Stage 4'' is the nuisance task's hardest group; ``Rare'' is the rarest pairwise mode. The original three configurations are reused, so this is not an independent rerun of them. Full standard deviations and selections are retained in the data archive.}
\label{tab:all_optimizers}
\begin{tabular}{lrrrrrr}\toprule
Configuration & Votes & Nuisance & Modes & Mixed & Stage 4 & Rare\\\midrule
SGDM & 67.19 & 63.72 & 85.86 & 74.45 & 50.24 & 55.58\\
Adam & 67.47 & 62.35 & 71.88 & 73.05 & 50.40 & 53.44\\
Muon & 62.19 & 63.06 & 86.81 & 74.96 & 49.86 & 67.56\\
Nesterov & 67.14 & 64.44 & 85.66 & 74.60 & 50.46 & 55.73\\
AdaBelief & 68.07 & 63.02 & 74.43 & 73.67 & 50.53 & 53.92\\
AdamW & 67.47 & 62.35 & 71.88 & 73.05 & 50.40 & 53.44\\
AdamW-EMA & 67.53 & 60.29 & 71.99 & 70.69 & 50.42 & 53.77\\
AdEMAMix & 67.44 & 62.22 & 71.81 & 73.10 & 50.40 & 54.65\\
ADOPT & 71.31 & 60.98 & 76.31 & 72.09 & 50.12 & 53.69\\
Cautious-AdamW & 70.52 & 62.52 & 73.11 & 73.74 & 50.11 & 53.62\\
Lion & 69.78 & 59.21 & 71.72 & 71.55 & 50.17 & 55.80\\
Muon-EMA & 63.35 & 60.74 & 87.56 & 74.41 & 50.12 & 68.58\\
NAdamW & 69.36 & 61.99 & 73.38 & 73.22 & 49.93 & 53.12\\
PAdam (0.25) & 66.85 & 62.99 & 71.79 & 74.19 & 50.42 & 53.86\\
RAdam & 71.44 & 62.05 & 71.30 & 73.13 & 50.15 & 53.83\\
RMSprop & 67.19 & 62.37 & 70.10 & 73.06 & 50.14 & 54.52\\
Schedule-free & 69.69 & 61.99 & 70.29 & 73.38 & 50.27 & 52.58\\
Shampoo & 65.97 & 63.61 & 83.98 & 74.73 & 50.37 & 54.43\\
Signum & 65.09 & 58.63 & 75.04 & 69.08 & 50.15 & 53.79\\
SOAP & 66.62 & 62.08 & 74.80 & 72.93 & 49.93 & 55.47\\
\bottomrule
\end{tabular}

\end{table}

\subsection{Source records and reproducibility scope}
The manuscript bundle preserves the five experiment reports, generator and runner snapshots, task and run configurations, selected seed records, diagnostic records, and derived figure/table data. Its \texttt{data/} directory includes the paired fourth-stage effects, cue diagnostics, rare-mode gaps, complete strict-order selections, boundary audit, optimizer matrix, and CIFAR-10 selections. \texttt{source\_data/} retains the underlying available records; \texttt{verification.json} records manuscript checks and source hashes. The derivation scripts reproduce the numerical tables without running training.

These records support the reported comparisons; they are not a claim that every original trajectory and model checkpoint is included in the manuscript archive. In particular, the twenty-configuration extension does not retain all stage-specific trajectories. The English source archive contains the files needed to compile the paper; the separate full archive additionally contains the Chinese comparison text and evidence records.

\end{document}